\pdfoutput=1

\documentclass[11pt]{article}

\usepackage{acl}

\usepackage{times}
\usepackage{latexsym}

\usepackage{microtype}

\usepackage{subcaption}
\usepackage{adjustbox}
\usepackage{graphicx}
\usepackage[utf8]{inputenc} 
\usepackage[T1]{fontenc}    
\usepackage{hyperref}       
\usepackage{url}            
\usepackage{booktabs}       
\usepackage{amsfonts}       
\usepackage{nicefrac}       
\usepackage{xcolor}         
\usepackage{epigraph}
\usepackage{algorithm}
\usepackage{algcompatible}
\usepackage{amsmath}
\usepackage{enumitem}
\usepackage[noend]{algpseudocode}
\usepackage{todonotes}

\algrenewcommand\algorithmicforall{\textbf{foreach}}
\algrenewcommand\algorithmicindent{.8em}

\usepackage{cleveref}
\Crefformat{section}{\S#2#1#3}
\Crefformat{subsection}{\S#2#1#3}
\Crefformat{subsubsection}{\S#2#1#3}
\Crefrangeformat{section}{\S\S#3#1#4 to~#5#2#6}
\Crefmultiformat{section}{\S\S#2#1#3}{ and~#2#1#3}{, #2#1#3}{ and~#2#1#3}
\Crefformat{figure}{#2Fig.~#1#3}
\Crefmultiformat{figure}{Figs.~#2#1#3}{ and~#2#1#3}{, #2#1#3}{ and~#2#1#3}
\Crefformat{table}{#2Tab.~#1#3}
\Crefmultiformat{table}{Tabs.~#2#1#3}{ and~#2#1#3}{, #2#1#3}{ and~#2#1#3}
\Crefformat{appendix}{#2Appx.~\S#1#3}
\Crefformat{algorithm}{Alg.~#2#1#3}
\Crefformat{equation}{#2Eq.~#1#3}

\title{\textsc{Devil's Advocate}: Anticipatory Reflection for LLM Agents}

\author{%
  Haoyu Wang\thanks{Work done during internship at Google DeepMind.} \\
  \normalsize{UPenn}\\
  \small{\texttt{why16gzl@seas.upenn.edu}} \\
  \And
  Tao Li \\
  \normalsize{Google DeepMind}\\
  \small{\texttt{tlinlp@google.com}} \\
  \And
  Zhiwei Deng \\
  \normalsize{Google DeepMind}\\
  \small{\texttt{zhiweideng@google.com}} \\
  \AND
  Dan Roth \\
  \normalsize{UPenn}\\
  \small{\texttt{danroth@seas.upenn.edu}} \\
  \And
  Yang Li \\
  \normalsize{Google DeepMind}\\
  \small{\texttt{liyang@google.com}} \\
}

\begin{document}

\maketitle

\begin{abstract}

In this work, we introduce a novel approach that equips LLM agents with introspection, enhancing consistency and adaptability in solving complex tasks. 
Our approach prompts LLM agents to decompose a given task into manageable subtasks (i.e., to make a plan), and to continuously introspect upon the suitability and results of their actions. 
We implement a three-fold introspective intervention: 
1) \textbf{anticipatory reflection} on potential failures and alternative remedy \textit{before} action execution, 
2) \textit{post}-action alignment with subtask objectives and backtracking with remedy to ensure \textbf{utmost effort in plan execution}, and 
3) comprehensive review upon plan completion for \textbf{future strategy refinement}. 
By deploying and experimenting with this methodology---a zero-shot approach---within WebArena for practical tasks in web environments, our agent demonstrates superior performance with a success rate of 23.5\% over existing zero-shot methods by 3.5\%. 
The experimental results suggest that our introspection-driven approach not only enhances the agent's ability to navigate unanticipated challenges through a robust mechanism of plan execution, but also improves efficiency by reducing the number of trials and plan revisions by 45\% needed to achieve a task.


\end{abstract}

\section{Introduction}

\vspace{-7pt}
\epigraph{Two roads diverged in a yellow wood,\newline
And sorry I could not travel both\newline
$\cdots$\newline
Then took the other, as just as fair,\newline
And having perhaps the better claim}{Robert Frost}
\vspace{-5pt}

\begin{figure}[t]
    \centering
    \includegraphics[width=0.9\linewidth]{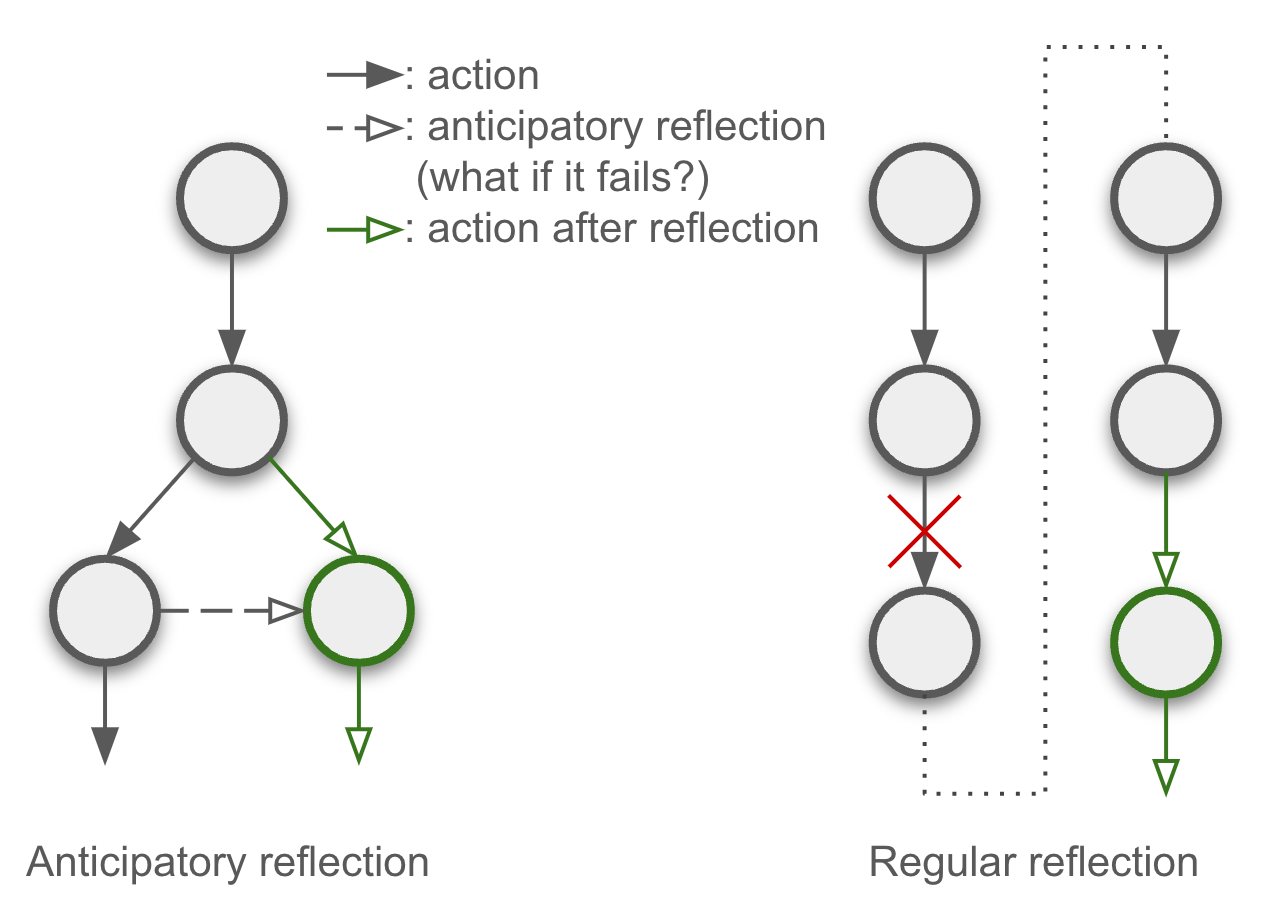}
    \caption{Conceptual difference between our anticipatory reflection and regular ones. 
    Circles denote states and arrows actions.
    At the branching level, our method does not only yield the next action, but also anticipates a potential error associated with it and plans for backups.
    In contrast, regular reflection performs trials sequentially, correcting one error for each pass.}
    \label{fig:reflection_diff}
\end{figure}

The enduring appeal of Frost's emblematic poem, ``The Road Not Taken,'' resides not just in its poetic elegance, but also in the profound lesson it imparts about decision-making. 
As we stand at the crossroads of a choice, it is a daunting challenge to assess probable outcomes and choose a course that best aligns with our objectives. 
This task becomes even more formidable when Large Language Model (LLM) agents \cite{huang2022inner, yao2023react, song2023llmplanner} have to navigate complex scenarios unfolding in real time, e.g., solving tasks in web environments \cite{liu2018reinforcement, yao2022webshop, deng2023mind2web, zhou2024webarena}, conducting simulated science experiments \cite{wang-etal-2022-scienceworld}, and solving embodied household tasks \cite{ALFWorld20}.

Indeed, LLM agent decision-making has witnessed enhancement by post-hoc reflection and correction \cite{shinn2023reflexion, song2024trial}, coupled with adaptive planning \cite{sun2023adaplanner, prasad2023adapt}, where the agents learn from past successes and failures while concurrently mapping out flexible strategies.
However, reflection usually works sequentially where only one hypothetical error can be corrected for each head-to-toe execution trajectory.
Considering that such reflection is a test-time strategy, it poses a great efficiency issue.
For instance, the agent could retry 10 times before concluding it still can not solve the task.
Furthermore, self-reflection involves frequent shifts in plans which, albeit a mere inconvenience for humans, can lead to disorientation for AI agents. 
This may produce confusion, a standstill, or even an infinite loop of failure, which substantiates the importance of \emph{thoroughly executing a set plan with utmost effort before resorting to a plan revision}. 
Therefore, this paper puts forward a methodology aimed at achieving an optimal balance between consistency and adaptability. This critical equilibrium mirrors the resilience and agility that is anticipated of a capable system that is prepared for curveballs but unwavering in the execution of its plan.
Fig.~\ref{fig:reflection_diff} highlight our design in comparison to existing reflection strategy.

In this paper, we introduce a novel approach that integrates introspection into the fabric of LLM agents. This approach enables agents to continuously reflect on their actions, thereby stimulating a learning process that dynamically optimizes exploration paths and enhances robust decision-making under uncertainty. Our introspective intervention focuses on three principal dimensions:
\begin{enumerate}[nosep]
    \item Anticipatory reflection before action execution (similar to a devil's advocate);
    \item Post-action evaluation and backtracking with remedy when necessary, to ensure the outcome aligns with subtask objectives;
    \item An extensive review upon plan completion to generate finer plans for subsequent trials.
\end{enumerate}

We implement this introspective methodology within WebArena \cite{zhou2024webarena}, a comprehensive web environment featuring 812 tasks in five scenarios: online shopping, e-commerce management, social discussion forums, maps, and software development platforms. Experimental results demonstrate that our approach, which is zero-shot, substantially outperforms state-of-the-art zero-shot methods while improving efficiency, paving the way for a new paradigm of intelligent systems that are more consistent, adaptable, and effective\footnote{Code to reproduce our results will be released.}.

\section{Related Works}
\label{related_work}
In this paper, we develop and expand upon several key themes within the realm of natural language processing, with a specific focus on the integration of action generation, planning, and reflection in the construction of LLM agents.

\paragraph{Action Generation} LLMs have been employed in tasks requiring decision-making or action generation and have proven useful as agent-controlling policies in embodied environments \cite{huang2022inner, huang2022language, driess2023palme, wang2023voyager, zhu2023ghost}. They have also demonstrated effectiveness in text-based environments \cite{liu2018reinforcement, ALFWorld20, liu2023agentbench}, where techniques like ReAct \cite{yao2023react} have shown notable benefits. Despite its success, ReAct's limitation lies in its inability to adjust to changes in the environment. Several improvements \cite{madaan2023selfrefine, shinn2023reflexion} have been proposed to counter these limitations, advocating for self-reflection to enhance decision-making and reasoning. However, these techniques primarily aim to improve single plans or trajectories without considering alternative actions, which could modify the plan in a wrong direction.

\paragraph{Position Bias Mitigation} While comparing answer choices is generally effective, large language models used for action generation are not without flaws. They can exhibit bias, especially towards the first (or sometimes second) answer they see, regardless of its quality. This is known as position bias \cite{zheng2023judging, wang2023large}. 
Our method mitigates this bias by asking follow-up questions that challenge its own answer.

\paragraph{Planning} Extensive research has explored the potential of LLMs in task planning \cite{dror-etal-2023-zero, prasad2023adapt, sun2023adaplanner, TaPA, guan2023leveraging, gur2024a}. The concept of decoupling planning and execution in formulating LLM agents has been validated through numerous paradigms such as ReWOO \cite{xu2023rewoo}, ADaPT \cite{prasad2023adapt}, Structured Self-Reflection \cite{li-etal-2023-zero}, and DEFS \cite{wang2023describe}. Nonetheless, these methods exhibit a deficiency in establishing a resilient mechanism for plan execution, with agents frequently revisiting and revising their plans following each instance of adverse environmental feedback, often due to inaccurately executed actions. Our approach, conversely, emphasizes executing a previously defined plan with unwavering effort before considering any modifications. This guarantees a more stable and consistent problem-solving process. To implement this, the factor of tree search becomes crucial for exploring the best solutions. Past approaches, including ToT \cite{yao2023tree}, RAP \cite{hao-etal-2023-reasoning}, LATS \cite{zhou2024language}, AdaPlanner \cite{sun2023adaplanner}, and ToolChain* \cite{zhuang2024toolchain}, have incorporated tree search techniques in identifying the optimal route to the desired solution. However, our approach distinguishes itself by engaging the LLM in preparing alternate solutions in anticipation of impending failures, ensuring more comprehensive consideration in action generation.

\paragraph{Reflection and Self-refinement} Reflection and refinement techniques have advanced significantly through works such as Reflexion \cite{shinn2023reflexion}, AdaPlanner \cite{sun2023adaplanner}, and AutoEval \cite{pan2024autonomous}. Our methodology further enhances this by incorporating an anticipatory reflection mechanism that operates before each action rather than performing post-hoc reflection after each complete trial. This approach simplifies exploration by expediting remedial action and reducing extensive backtracking and serial plan revisions, thereby improving the overall efficiency.

\section{Method}
\label{method}

Given a task \( \mathcal{T} \) and an environment \( \mathcal{E} \) with which the LLM agent \( G \) interacts, our objective is to enable the agent to systematically and adaptively complete the task through introspective methods.
We first present how we decompose the task and generate action regarding each state in the environment in \cref{subsec:decom} and \cref{subsec:state}. Then we introduce the introspection mechanism in \cref{subsec:intros}.

\subsection{Task Decomposition and Planning}
\label{subsec:decom}

\begin{figure}
    \centering
    \includegraphics[width=1\linewidth]{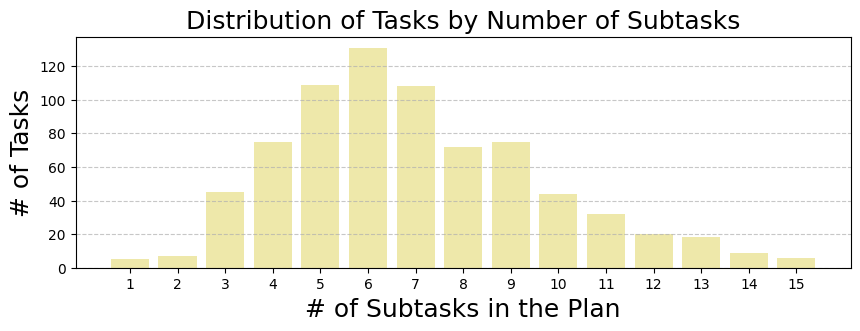}
    \caption{Distribution of WebArena tasks based on the number of subtasks within each task.
    The number of subtasks has a majority within 4-9 with a long tail distribution.}
    \label{fig:dist}
\end{figure}

The first step involves decomposing the task \( \mathcal{T} \) into subtasks in a sequential manner, forming a plan. This decomposition is achieved through an LLM generation process. Let \( G_{\text{plan}} \) denote the agent's plan generation function, prompted by the task \( \mathcal{T} \), description of the initial state \( S_0 \), and any experience from past trials, i.e., history \( \mathcal{H} \):
\begin{align}
    \mathcal{P} \sim G_{\text{plan}}(\mathcal{T}, S_0, \mathcal{H}).
\end{align}
Here, the plan \( \mathcal{P} \) is parsed into a sequence of ordered subtasks:
\begin{align}
    \mathcal{P} = (\tau_1, \tau_2, \ldots, \tau_N),
\end{align}
where \( \tau_i \) represents the \( i \)-th subtask in the plan, and \( N \) is the number of subtasks. For instance, Fig. \ref{fig:plan} shows a plan with 5 subtasks for solving a task in WebArena.
The distribution of WebArena tasks based on the number of subtasks within each task is illustrated in Fig. \ref{fig:dist}. 
This also reflects the difficulty of the tasks in WebArena, where most tasks take 4-9 steps to complete.
\begin{figure}[t]
    \centering
    \fbox{
    \parbox{0.95\linewidth}{
    \textbf{Plan for task: \textit{What is the color configuration of the picture frame I bought in Nov 2022}:}\\
    1. Click on the `My Account' link to access your account details. \\
    2. Click on the `Order History' link to view your past orders.\\
    3. Scroll down the page until you find the order from November 2022.\\
    4. Click on the order details link for the order from November 2022.\\
    5. Scroll down to the product details section to find the color configuration of the picture frame.
    }}
    \caption{An example plan with 5 subtasks, generated by GPT-4.
    Subtasks are generated based on the first observation $\mathcal{S}_0$ and prior knowledge about web operation.}
    \label{fig:plan}
\end{figure}

\begin{algorithm*}[t]
\small
\caption{Introspective Agent}
\label{algo}
Input: task $\mathcal{T};$ initial observation $S_\text{initial};$ environment $\mathcal{E};$\\
Initialization: time $t = 0;$ state $S_t = S_\text{initial};$ action $a_t=\emptyset;$ plan $\mathcal{P}=\emptyset;$ subtask $\tau = \emptyset;$ history $\mathcal{H}=\emptyset;$
\begin{algorithmic}[1]

    \While {$\neg G_{\text{completed}}(\mathcal{T}, \cdot)$}
        \State $\mathcal{P}\sim G_{\text{plan}}(\mathcal{T}, S_t, \mathcal{H});$ \Comment{Plan Revision}
        \State $\mathrm{Stack} = [(S_t, a_t, \tau)];$
        \While{$\mathrm{Stack}$}
            \State $(S_t', a_t, \tau) = \mathrm{Stack}.\text{pop}()$
            \If {$S_t \neq S_t'$}
                $\text{go\_back}(S_t'); S_t = S_t';$ \Comment{Backtracking}
            \EndIf
            \If{$\tau$ \textbf{is} $\emptyset$}
                 $\mathcal{C}_{\tau} = 1; \tau = \mathcal{P}.\text{next}();$
            \Else
                 \textbf{ }$S_{t+1} = \mathcal{E}(a_t); \mathcal{H}.\text{add}(G_{\text{describe}}(S_t, a_t, S_{t+1}));$ \Comment{Grounding}
                 \State $ \mathcal{C}_{\tau} \sim G_{\text{align}}(S_t, a_t, S_{t+1}, \tau);$ \Comment{Alignment with Subtask Objective}
                \If{$\mathcal{C}_{\tau}$}
                    \If{$G_{\text{completed}}(\mathcal{T}, S_{t+1})$}
                    Finished; \Comment{Early Stop}
                    \EndIf
                    \If{$G_{\text{completed}}(\tau, S_{t+1})$}
                    $\tau = \mathcal{P}.\text{next}()$; \Comment{Next Subtask}
                    \EndIf
                \EndIf
            \EndIf
            \State $t\texttt{++};$
            \If{$\mathcal{C}_{\tau}$}
                 $a_{t} \sim G_{\text{action}}(\tau, S_{t});$
                \For{$r = 1$ \textbf{to} $R$} 
                    \State $a_{t}^{(r)} \sim G_{\text{remedy}}(\tau, S_{t}, a_{t});$ \Comment{Anticipatory Reflection}
                    \State $\mathrm{Stack}.\text{push}((S_{t}, a_{t}^{(r)}, \tau));$
                \EndFor
                \State $ \mathrm{Stack}.\text{push}((S_{t}, a_{t}, \tau));$ \Comment{Placing $a_t$ at the top of $\mathrm{Stack}$}
            \EndIf

        \EndWhile
        
    \EndWhile
\end{algorithmic}
\end{algorithm*}

\subsection{State and Action Representation}
\label{subsec:state}

Let \( S_t \in \mathcal{S} \) denote the current state of the environment at time \( t \), where \( \mathcal{S} \) is the set of all possible states. 
From state \( S_t \), let \( a_t \in \mathcal{A} \) denote the next action taken by the agent, where \( \mathcal{A} \) is the set of all possible actions.
The next action is generated based on the the specific subtask \( \tau_i \) being addressed, current state \( S_t \), and action history \( \mathcal{H}_{t-1} \): 
\begin{align}
a_t \sim G_{\text{action}}(\tau_i, S_t, \mathcal{H}_{t-1}),
\end{align}
where \( G_{\text{action}} \) denotes the agent's action generation function.
Let \( \mathcal{H}_{t} \) denote the history of actions taken up to time \( t \):
\begin{align}
    \mathcal{H}_{t} = \{\hat{a}_1, \hat{a}_2, \ldots, \hat{a}_{t}\},
\end{align}
where \( \hat{a}_{t} \) is a textual description of action \( a_t \), along with useful information learned from this action execution, generated with function \( G_{\text{describe}} \). 
The history would later be used to answer questions in the task or to revise the agent's plan.
\( G_{\text{describe}} \) accepts as input the state before the action, the action itself, the state after the action:
\begin{align}
    \hat{a}_{t} \sim G_{\text{describe}}(S_{t}, a_{t}, S_{t+1}).
\end{align}
When the state observation is too long to fit in the context window of an LLM, the state is first summarized by the LLM into a shorter description before being fed to \( G_{\text{describe}} \) (e.g., this operation is commonly needed for solving web navigation tasks on content management platforms).
Note that a subtask can involve several actions, and thus \(i\) does not necessarily equal to \(t\).
Given the possibility that the task can be finished at some time \(t\) before the completion of all subtasks, whenever the agent arrives at a new state, we ask the agent to check two things: whether the subtask is finished \(\mathcal{C}_{\tau_i} \in (0, 1)\)\footnote{When the agent determines that a subtask is non-essential to solving the task, we also set \(\mathcal{C}_{\tau_i} = 1\).}, and whether the task is finished \(\mathcal{C}_{\mathcal{T}} \in (0, 1)\):
\begin{align}
    \mathcal{C}_{\tau_i} &\sim G_{\text{completed}}(\tau_i, S_{t+1}, \mathcal{H}_t), \\ \mathcal{C}_{\mathcal{T}} &\sim G_{\text{completed}}(\mathcal{T}, S_{t+1}, \mathcal{H}_t),
\end{align}
where \(G_{\text{completed}}\) denotes the function for checking whether an objective is fulfilled. If \(\mathcal{C}_{\tau_i} = 1\), the agent moves on to solve the next subtask \(\tau_{i+1}\); 
whereas when the agent determines \(\mathcal{C}_{\mathcal{T}} = 1\), it finishes the current trial regardless of whether the plan \( \mathcal{P} \) is finished.

\subsection{Introspective Mechanisms}
\label{subsec:intros}
The sequential action generation above can potentially execute the plan and solve the task already.
Nevertheless, without proper introspection and adaptation, the agent might be stuck at a certain unsolvable subtask or go into a loop of failure when unexpected problems emerge. Thus, we introduce three introspective mechanisms to enhance our LLM agent's problem-solving ability below.

\subsubsection{Anticipatory Reflection (\textsc{Devil's Advocate})}
\begin{figure}[t]
    \centering
    \includegraphics[width=0.98\linewidth]{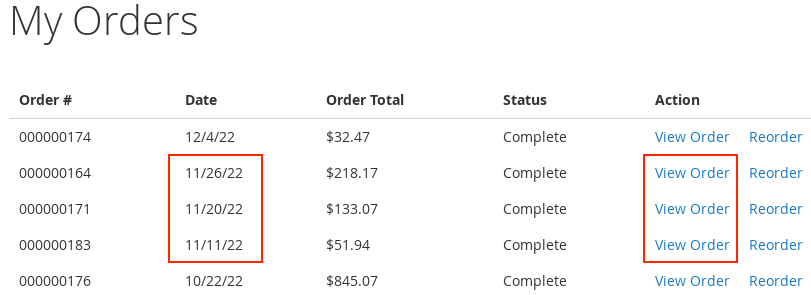}
    \caption{Screen observation at one step in solving 
    the subtask: \textit{Click on the order details link for the order from November 2022}. The agent might decide to click (\(a_t\)) on the ``View Order'' button of \textbf{any one of the three Nov 2022 orders} to see if a picture frame was purchased in that order, and it is highly probable that backtracking is needed to view the details of the other two orders (if the first chosen is not a picture frame). In our proposed approach, the other two alternative clicking actions \([a_t^{1}, a_t^{2}]\) would be pushed to stack before the agent executes action \(a_t\).}
    \label{fig:anticipatory_reflection}
\end{figure}

The first layer of introspection occurs before each action execution. 
The agent anticipates potential failures and comes up with $R$ alternative remedies \([ a_t^{1}, a_t^{2}, \cdots, a_t^{R} ]\). Each remedy action is generated by prompting the LLM with a follow-up question: 
\begin{itemize}
    \item \textsl{"If your answer above is not correct, instead, the next action should be:"}
\end{itemize}
We use \(G_{\text{remedy}}\) to denote the generation of remedy actions, which accepts as input the subtask \( \tau_i \), the current state \( S_t \), the action history \(\mathcal{H}_{t-1}\), and the LLM predicted next action \(a_t\) at first attempt:
\begin{align}
a_t^{r} &\sim G_{\text{remedy}}(\tau_i, S_t, \mathcal{H}_{t-1}, a_t).
\end{align}
If later found necessary, the agent can go back to state \(S_t\) to modify the original action \(a_t\) to try the remedy action \(a_t^{r}\) to ensure a smooth plan execution.
For example, in Fig. \ref{fig:anticipatory_reflection}, we show a state observation where all three clicking actions align with the objective of the current subtask. 
The execution of any of these actions would complete the subtask; yet the agent might need to return to this state if it later determines that the action predicted at first attempt was incorrect\footnote{The action generated at first attempt still gets the highest priority, i.e., \(a_t\) is the last one to be pushed to the stack so it can be popped and executed first (see line 18 in Alg. \ref{algo}).}.

\subsubsection{Post-action Evaluation and Backtracking}

The second introspective mechanism kicks in after the execution of each action. Here, the agent evaluates whether the action and the resulting state align with the subtask objective. This introspective function, denoted as \(G_{\text{align}}\), is motivated by the state before the action \(S_t\), the action \(a_t\), the resulting state \(S_{t+1}\), the current subtask \(\tau_i\):
\begin{align}
\theta_t &\sim G_{\text{align}}(S_t, a_t, S_{t+1}, \tau_i).
\end{align}
Here \(\theta_t \in (0, 1)\) denotes the evaluation score reflecting how well the state \(S_{t+1}\) aligns with the subtask objective \(\tau_i\).
It is a binary signal indicating whether the agent needs to stop and backtrack to some previous state and take an alternative action \(a_k^{r}, k \le t\), if the execution of \(a_t\) does not meet the objective of the current subtask.
In our experiments with web environments, the URL of the webpage is a useful information recorded as part of \(S_t\). 
When backtracking, we can easily navigate back to the URL. However, the element information on the URL might differ from the state we first encountered upon arriving at that page. 
To address this, we prompt the LLM to map the recorded element in the action to the new element with which we want to interact, if necessary.

\subsubsection{Plan Revision}

The third introspective mechanism occurs upon plan failure, i.e., when the stack is empty and \(\mathcal{C}_{\mathcal{T}} = 0\). 
Now the agent performs a thorough review of the actions executed and the notes taken, and refines its future plan based on identified problems: 
\begin{align}
\mathcal{P}_{\text{new}} &\sim G_{\text{plan}}(\mathcal{T}, S_0, \mathcal{H}_t).
\end{align}

Here, \(\mathcal{P}_{\text{new}}\) is the new plan after reflecting on the past failed trials. 
The agent then re-enters the plan execution phase and starts a new episode.

Through these three layers of introspection, our agent is more capable of navigating the complexities of unforeseen circumstances and addressing tasks, bringing us a significant stride closer to achieving truly autonomous, adaptable, and intelligent systems.
By structuring the problem in this manner, we have established a clear framework for enabling LLM agents to perform tasks autonomously and adaptively through introspection. Alg. \ref{algo} shows a pseudo code of our approach.

\begin{figure*}[t]
    \centering
    \includegraphics[clip,trim={2.4cm 0.0cm 2.2cm 0.0cm},width=0.9\linewidth]{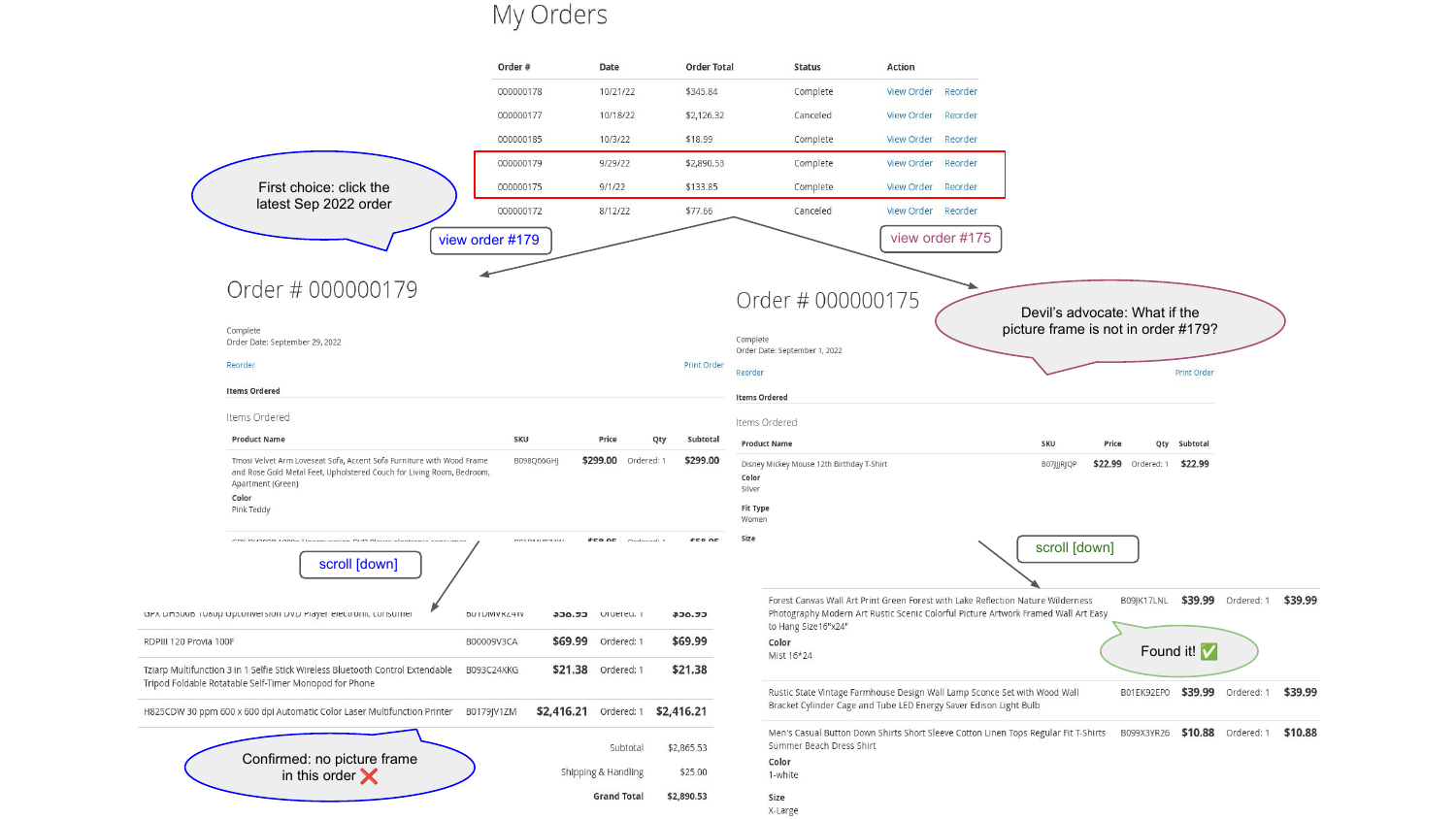}
    \caption{Decision making process of our agent in solving the task: \textit{What is the color configuration of the picture frame that I bought in Sep 2022?} Before execution of the predicted action, the agent asks a follow-up question to itself regarding its decision: \textit{what if the picture frame is not in order \#179? what should be the alternative remedy?} And after finding out that order \#179 contains no picture frame at all, the agent backtracks to the previous state to view order \#175 and continue.}
    \label{fig:DP}
\end{figure*}

\section{Experiments}

In this section, we demonstrate how introspection enhances consistency and adaptability of LLM agents in solving complex tasks in web environments.
We first introduce the experimental setup for evaluation (\cref{subsec:exp_setup}), followed by evaluation results (\cref{subsec:results}). Detailed error analysis is provided in \cref{sec:error}, which highlights the directions for future endeavor.
\subsection{Experimental Setup}
\label{subsec:exp_setup}
\paragraph{Live Environments} We evaluate our proposed method in the simulated web environments of WebArena \cite{zhou2024webarena}, a dataset of human-annotated web browsing tasks designed to evaluate the ability of LLMs to perform complex, real-world actions on the internet\footnote{Webarena (\url{https://webarena.dev}) is licensed under a Creative Commons Attribution-ShareAlike 4.0 International License.}.
The 812 tasks in WebArena involve five websites: an online shopping website, a software development website, a social forum platform, a map, and an e-commerce management platform; and these tasks can be categorized into three classes: information seeking tasks, site navigation and content \& config tasks, and unachievable tasks.
Though WebArena provides visual observation (screenshots), in this work we use the text observation only.
The observation at each step is the accessibility tree of the webpage, and the elements in the accessibility tree are all within the current viewport of a 1280$\times$720 screen.
The action space of our LLM agent includes actions that interact with environment: \textsl{click}, \textsl{type}, \textsl{scroll}, \textsl{goto}, \textsl{go\_back}, \textsl{go\_forward}, and also a \textsl{note\_down} action that takes down useful snippet/summary for answering information-seeking questions. 

\paragraph{Baselines} We employ \texttt{gpt-4-0613}\footnote{\url{https://platform.openai.com/docs/models/gpt-4-turbo-and-gpt-4}} \cite{achiam2023gpt} with a context window of 8k tokens to build the agents and compare our method with three other agent construction strategies: planning and sequential decision making (Plan + Act w/o reflexion), similar to ReWOO \cite{xu2023rewoo}; planning and sequential decision making with reflection (Plan + Act), similar to AdaPlanner \cite{sun2023adaplanner}; and tree search based planning, similar to LATS \cite{zhou2024language}, but with reflection.
In all methods, we set the upper limit on the number of actions to 30, i.e., after the agent executes 30 actions for a given task, it has to stop.
In all three methods, we adopt the same prompts for action generation \(G_\text{action}\), plan generation \(G_{\text{plan}}\), and evaluator \(G_{\text{align}}\) and \(G_{\text{completed}}\) to ensure a fair comparison\footnote{Detailed prompts are shown in the Appendix.}.
In our experiments, we set the LLM  temperature to 1.0 and max\_tokens to 512, and keep all other parameters as default.

\paragraph{Metrics}
We follow the evaluation metric ``Success Rate'' in \cite{zhou2024webarena}, and count the number of actions per trial and the number of plan revisions per task.
To determine whether a task is successfully completed, the \texttt{exact\_match} metric is used for some site navigation and information seeking tasks. 
However, this can sometimes be overly stringent. 
For instance, consider the URLs below that display the same content (under `electronics', the category id of `headphones' is  60).
In fact, both of them point to exactly the same webpage.
However, when evaluating for task completion, only the one
that exactly matches a predefined finish URL is considered correct\footnote{In WebArena, only the first URL link is used as the ground truth thus agent that reaches the second URL is judged as task incomplete.}. 
To address this issue, we manually review the evaluation process and correct such misjudgements in our results\footnote{Our manual correction will also be released together with our code.}. 
\begin{itemize}[nosep]
\item \small \url{http://localhost:7770/electronics/headphones.html}
\item \small \url{http://localhost:7770/electronics.html?cat=60}
\end{itemize}



\subsection{Results}
\label{subsec:results}

\begin{figure}
    \centering
    \includegraphics[width=1\linewidth]{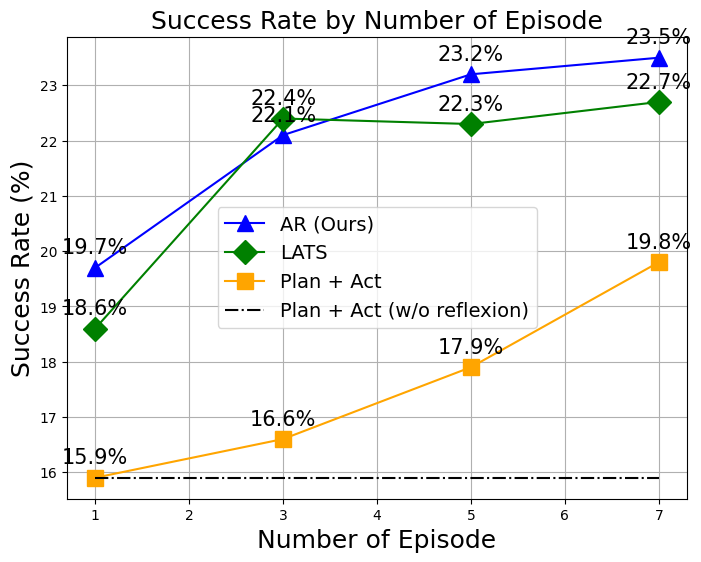}
    \caption{Results of different agent construction strategies on WebArena. AR is short for our method, anticipatory reflection; LATS represents our in-house implementation of the approach proposed by  \citet{zhou2024language}; Plan + Act is a method of decomposition of task and execution of each subtask, similar to ReWOO \cite{xu2023rewoo}. All three methods are equipped with plan revision (post-failure reflection).} 
    \label{fig:reflexion}
\end{figure}
The experimental results, depicted in Fig. \ref{fig:reflexion}, demonstrate the efficacy of our introspection-driven approach in enhancing the consistency and adaptability of LLM agents in web environments. We compare the success rates of various agent construction strategies across multiple episodes. Our method, anticipatory reflection (AR), consistently outperforms the others, achieving a success rate of 23.5\% after seven episodes, closely followed by LATS with 22.7\%. In contrast, the Plan + Act method shows gradual improvement, reaching 19.8\%, but remains significantly lower than the tree-search-based AR and LATS methods.
Taking a closer look at the performance curve of LATS, there is an inconsistent pattern as success rate even drops at round 5.
This is likely due to the homogeneous generated actions through direct sampling.
In comparison, AR benefits from the ``devil's advocate'' approach, enabling more thorough planning and execution due to introspective follow-up questions.
This trend underscores the importance of incorporating introspection mechanisms for both plan execution and revision, highlighting their critical role in enhancing consistency and efficiency.

\begin{table}[t]
    \centering\small
    \setlength{\tabcolsep}{4pt}
    \begin{tabular}{l|cc|c} %
    \toprule
        & \multicolumn{2}{c|}{\# of Actions} & \# of Plan Revisions \\ %
        & First Trial & Last Trial & \\ %
        \midrule
        Plan+Act & 4.01 & 4.47 & 2.03\\
        LATS & 6.08 & 6.45 & 1.16\\
        \midrule
        AR & 6.39 & 7.07 & 0.64\\
        \bottomrule
    \end{tabular}
    \caption{
    Statistics of the trajectory of different agents solving tasks on WebArena. We report the number of actions in the first and last trial, and also the number of plan revisions, i.e., trials.
    }
    \label{tab:eval_acc}
\end{table}

Further insights can be gleaned from \Cref{tab:eval_acc}, which compares the average number of actions in the first and last trials across different methods. Our AR method shows an increase in the average number of actions from 6.39 in the first trial to 7.07 in the last trial, indicating a robust learning and adaptation process. In comparison, the average number of actions in the first trial of the Plan+Act method is only 4.01, suggesting that it stops at an early stage without completing full plan execution. Thus, our method effectively leverages a greater number of actions to achieve better outcomes, thereby reducing the number of plan revisions by 45\% and improving overall efficiency.

\section{Error Analyses} 
\label{sec:error}
The subsequent sections shed light on an analysis of errors we observed from the agent's behavior when executing tasks. 
Two key areas have been identified for detailed discussion: an agent's occasional inability to fully learn from past failures, and inefficiencies in solving specific kinds of tasks due to a sequential planning scheme. 
\subsection{Agent Only Takes Partial Lesson from Past Failures}
\begin{figure} 
    \centering
    \includegraphics[width=\linewidth]{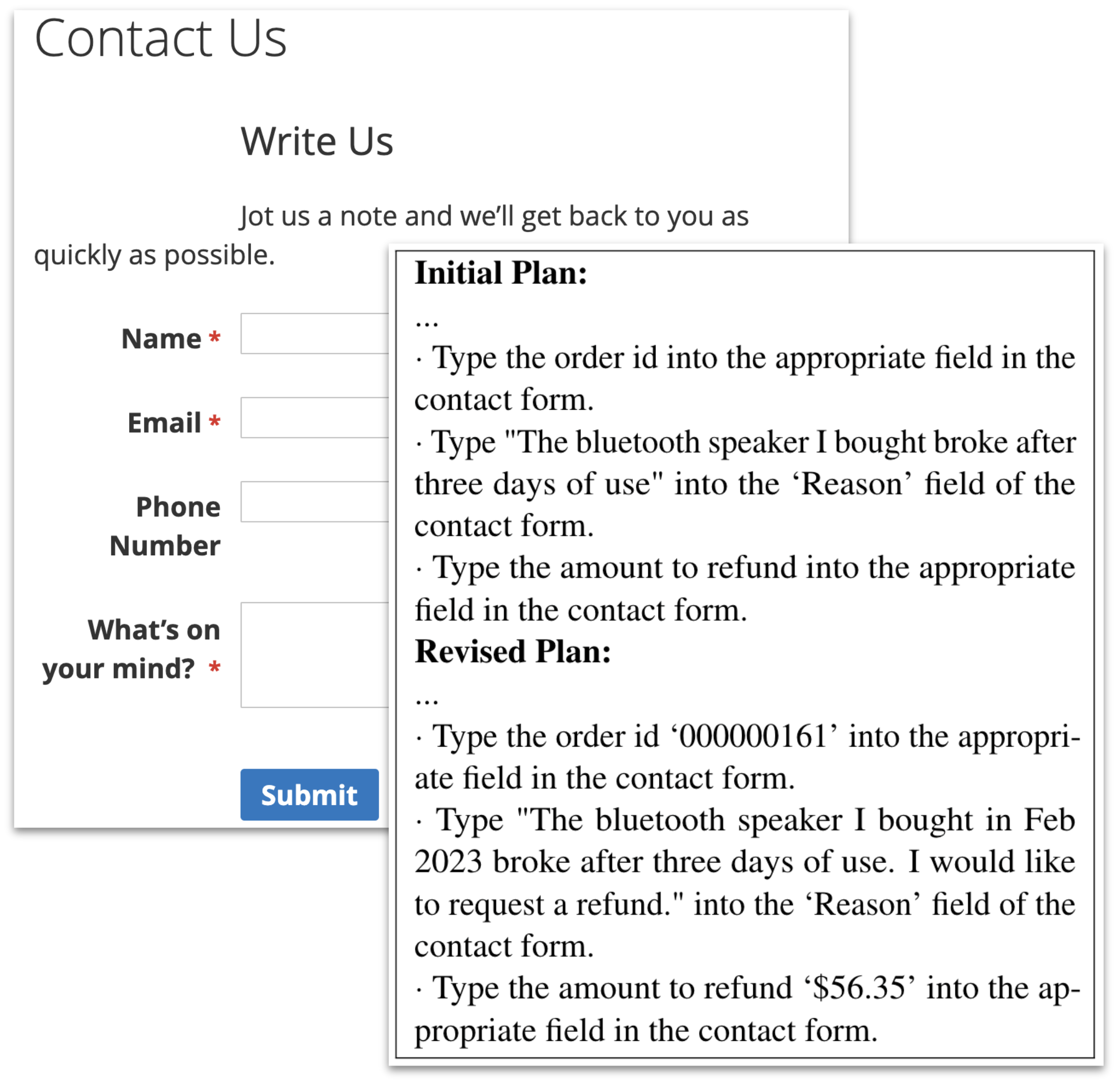}
    \caption{Screen observation at the last step to solve the task: \textit{Draft a refund message via their "contact us" form for the bluetooth speaker I bought Feb 2023. It broke after three days of use. The shop requires the order id, the reason and the amount to refund in the message. Don't submit yet.}}
    \label{fig:contact}
\end{figure}
One category of common errors we notice is that the agent is not taking full lesson from past failure when generating a new plan.
As illustrated in Fig. \ref{fig:contact}, the agent is at the final step of drafting a refund message for a Bluetooth speaker, after a series of steps taken to seek information for the order.
From the screen, we know that the agent should consolidate all the information gathered from previous steps and type one piece of text into the (only) box titled \textsl{``What's on your mind?''}.
However, as can be seen from the plans at the lower right corner in Fig. \ref{fig:contact}, while some improvements were made by adding the date of purchase and a more detailed explanation in the revised plan, the agent still failed to optimize the input process, repeating the typing actions separately for fields that do not exist. 
This inefficiency in the agent's behavior showcases the need for either an LLM with stronger reasoning ability or a better mechanism to solicit more comprehensive and accurate reflection.

\subsection{Sequential Planning is Not Enough}
In our analysis, we observed a recurrent error pertaining to the design of the agent's planning process. The proposed methodology structures a plan as a sequence of tasks that are executed in a specific order. Though it is effective in a decent amount of use cases, it seems to falter when faced with tasks necessitating more sophisticated logic. Specifically, tasks that mandate implementing a reusable function encapsulating several actions and employing a loop construct tend to challenge the model's current configuration. For example: \\
\begin{itemize}[nosep]
    \item \small \textsl{List out reviewers, if exist, who mention about average print quality.}
    \item \small \textsl{Give me the SKU of the products that have 1-3 units left.}
    \item \small \textsl{Like all submissions created by CameronKelsey in subreddit earthporn.}\\
\end{itemize}

Performing such tasks is analogous to executing SQL commands without a direct query API, but instead, in a realistic environment.
The ability to process these tasks effectively would necessitate the incorporation of additional cognitive constructs into the planning model—e.g., memory, loops, repetitive actions, or encapsulation of a group of actions into callable functions. 
Though taking notes can help the agent eliminate wrong choices, these systemic extensions would add crucial capabilities to the web agent, significantly enhancing its navigation and problem-solving competence in realistic web environments.
Moreover, while the current agent can succeed in the limited search space of simple tasks, it often struggles to review and introspect upon more
descriptive tasks that require dynamic problem-solving. 
By addressing these limitations in future work, i.e., effectively converting textual description of a plan into robust execution of callable functions and loops, we believe that the reasoning capability of our agent can be substantially improved, leading to better outcomes in understanding and solving tasks that involve dynamic cognition in web environments.
\section{Conclusions}

In this work, we introduce a novel introspective methodology that significantly enhances the problem-solving capabilities of LLMs in complex environments, as demonstrated through comprehensive evaluations in the WebArena setting. Our approach strategically decomposes tasks into actionable subtasks and incorporates a three-tiered introspection process, which includes anticipatory reflection, robust post-action evaluation, and episode-level plan revision. This setup not only allows LLM agents to adapt their strategies in real time but also fosters long-term learning, reducing the need for frequent interventions as experience accumulates.
The application of our introspective agent design in the WebArena benchmark demonstrates substantial performance gain (3.5\%) over state-of-the-art zero-shot approach, along with stable performance curve with increasing number of rounds.
Such benefits are accompanied by almost halving the number of plan revisions (45\%) during error handling. 
In summary, by enabling LLM agents to proactively contemplate potential failures, evaluate actions post-execution, and continuously refine their strategy based on experiential insights, our approach equips AI systems with a human-like strategic thinking capability.

\section*{Broader Impact}
Looking forward, the integration of multi-modal data inputs could further enhance the contextual understanding and decision-making accuracy of these agents. The principles and findings from our approach provide a robust foundation for future research in AI, particularly in aspects of autonomous decision-making, learning efficiency, and adaptability. As AI continues to integrate into diverse aspects of decision-making, embedding introspective capabilities will be essential to ensure these systems operate not only with precision but with an understanding akin to strategic human cognition.

\section*{Ethics Statement}
As the capabilities of LLM agents enhance and their deployment in real-world applications increases, it is crucial to address potential ethical concerns, particularly regarding data privacy, bias, and transparency. Our work focuses on improving agent introspection to enhance task performance and decision-making explanations, aiming to develop more transparent and trustworthy AI systems. We emphasize the importance of human oversight to monitor and mitigate unforeseen consequences and encourage the responsible use of this technology for societal benefit. By promoting continuous evaluation and fair practices, we seek to minimize biases and ensure that the deployment of these agents does not exacerbate social inequalities. Furthermore, we are committed to optimizing computational resources to reduce the environmental impact, advocating for sustainable AI practices.
\section*{Limitations}
Despite substantial progress made with our current design, limitations persist that inhibit optimal performance. Notably, the agent lacks a full learning mechanism to capitalize on past failures when generating a new plan, resulting in inefficient execution and recurring mistakes. Furthermore, while the sequential planning approach is effective for simpler tasks, it falls short for more sophisticated operations, such as those requiring encapsulated actions or loop constructs. Additionally, the agent struggles with tasks that expand beyond a simple search space, suggesting obstacles in handling dynamic problem-solving. 
Last but not least, our agent needs significant amounts of LLM generation (i.e., API calling), consequently requiring substantial time and computational resources, which dents its efficiency. Therefore, future work needs to concentrate on improving the agent's ability to fully learn from prior shortcomings, adapt to handle complex tasks, enhance dynamic problem-solving capabilities, and optimize time and resource utilization with more efficient LLM calling.

\bibliographystyle{acl_natbib}

\begin{thebibliography}{34}
\expandafter\ifx\csname natexlab\endcsname\relax\def\natexlab#1{#1}\fi

\bibitem[{Achiam et~al.(2023)Achiam, Adler, Agarwal, Ahmad, Akkaya, Aleman,
  Almeida, Altenschmidt, Altman, Anadkat et~al.}]{achiam2023gpt}
Josh Achiam, Steven Adler, Sandhini Agarwal, Lama Ahmad, Ilge Akkaya,
  Florencia~Leoni Aleman, Diogo Almeida, Janko Altenschmidt, Sam Altman,
  Shyamal Anadkat, et~al. 2023.
\newblock Gpt-4 technical report.
\newblock \emph{arXiv preprint arXiv:2303.08774}.

\bibitem[{Deng et~al.(2023)Deng, Gu, Zheng, Chen, Stevens, Wang, Sun, and
  Su}]{deng2023mind2web}
Xiang Deng, Yu~Gu, Boyuan Zheng, Shijie Chen, Samuel Stevens, Boshi Wang, Huan
  Sun, and Yu~Su. 2023.
\newblock \href {http://arxiv.org/abs/2306.06070} {Mind2Web: Towards a
  Generalist Agent for the Web}.

\bibitem[{Driess et~al.(2023)Driess, Xia, Sajjadi, Lynch, Chowdhery, Ichter,
  Wahid, Tompson, Vuong, Yu, Huang, Chebotar, Sermanet, Duckworth, Levine,
  Vanhoucke, Hausman, Toussaint, Greff, Zeng, Mordatch, and
  Florence}]{driess2023palme}
Danny Driess, Fei Xia, Mehdi S.~M. Sajjadi, Corey Lynch, Aakanksha Chowdhery,
  Brian Ichter, Ayzaan Wahid, Jonathan Tompson, Quan Vuong, Tianhe Yu, Wenlong
  Huang, Yevgen Chebotar, Pierre Sermanet, Daniel Duckworth, Sergey Levine,
  Vincent Vanhoucke, Karol Hausman, Marc Toussaint, Klaus Greff, Andy Zeng,
  Igor Mordatch, and Pete Florence. 2023.
\newblock PaLM-E: An Embodied Multimodal Language Model.
\newblock In \emph{arXiv preprint arXiv:2303.03378}.

\bibitem[{Dror et~al.(2023)Dror, Wang, and Roth}]{dror-etal-2023-zero}
Rotem Dror, Haoyu Wang, and Dan Roth. 2023.
\newblock \href {https://doi.org/10.18653/v1/2023.findings-eacl.53} {Zero-Shot
  On-the-Fly Event Schema Induction}.
\newblock In \emph{Findings of the Association for Computational Linguistics:
  EACL 2023}, pages 705--725, Dubrovnik, Croatia. Association for Computational
  Linguistics.

\bibitem[{Guan et~al.(2023)Guan, Valmeekam, Sreedharan, and
  Kambhampati}]{guan2023leveraging}
Lin Guan, Karthik Valmeekam, Sarath Sreedharan, and Subbarao Kambhampati. 2023.
\newblock \href {http://arxiv.org/abs/2305.14909} {Leveraging Pre-trained Large
  Language Models to Construct and Utilize World Models for Model-based Task
  Planning}.

\bibitem[{Gur et~al.(2024)Gur, Furuta, Huang, Safdari, Matsuo, Eck, and
  Faust}]{gur2024a}
Izzeddin Gur, Hiroki Furuta, Austin~V Huang, Mustafa Safdari, Yutaka Matsuo,
  Douglas Eck, and Aleksandra Faust. 2024.
\newblock \href {https://openreview.net/forum?id=9JQtrumvg8} {A Real-World
  WebAgent with Planning, Long Context Understanding, and Program Synthesis}.
\newblock In \emph{The Twelfth International Conference on Learning
  Representations}.

\bibitem[{Hao et~al.(2023)Hao, Gu, Ma, Hong, Wang, Wang, and
  Hu}]{hao-etal-2023-reasoning}
Shibo Hao, Yi~Gu, Haodi Ma, Joshua Hong, Zhen Wang, Daisy Wang, and Zhiting Hu.
  2023.
\newblock \href {https://doi.org/10.18653/v1/2023.emnlp-main.507} {Reasoning
  with Language Model is Planning with World Model}.
\newblock In \emph{Proceedings of the 2023 Conference on Empirical Methods in
  Natural Language Processing}, pages 8154--8173, Singapore. Association for
  Computational Linguistics.

\bibitem[{Huang et~al.(2022{\natexlab{a}})Huang, Abbeel, Pathak, and
  Mordatch}]{huang2022language}
Wenlong Huang, Pieter Abbeel, Deepak Pathak, and Igor Mordatch.
  2022{\natexlab{a}}.
\newblock Language Models as Zero-Shot Planners: Extracting Actionable
  Knowledge for Embodied Agents.
\newblock \emph{arXiv preprint arXiv:2201.07207}.

\bibitem[{Huang et~al.(2022{\natexlab{b}})Huang, Xia, Xiao, Chan, Liang,
  Florence, Zeng, Tompson, Mordatch, Chebotar, Sermanet, Brown, Jackson, Luu,
  Levine, Hausman, and Ichter}]{huang2022inner}
Wenlong Huang, Fei Xia, Ted Xiao, Harris Chan, Jacky Liang, Pete Florence, Andy
  Zeng, Jonathan Tompson, Igor Mordatch, Yevgen Chebotar, Pierre Sermanet, Noah
  Brown, Tomas Jackson, Linda Luu, Sergey Levine, Karol Hausman, and Brian
  Ichter. 2022{\natexlab{b}}.
\newblock Inner Monologue: Embodied Reasoning through Planning with Language
  Models.
\newblock In \emph{arXiv preprint arXiv:2207.05608}.

\bibitem[{Li et~al.(2023)Li, Li, Deng, Wang, and Li}]{li-etal-2023-zero}
Tao Li, Gang Li, Zhiwei Deng, Bryan Wang, and Yang Li. 2023.
\newblock \href {https://doi.org/10.18653/v1/2023.findings-emnlp.753} {A
  Zero-Shot Language Agent for Computer Control with Structured Reflection}.
\newblock In \emph{Findings of the Association for Computational Linguistics:
  EMNLP 2023}, pages 11261--11274, Singapore. Association for Computational
  Linguistics.

\bibitem[{Liu et~al.(2018)Liu, Guu, Pasupat, Shi, and
  Liang}]{liu2018reinforcement}
Evan~Zheran Liu, Kelvin Guu, Panupong Pasupat, Tianlin Shi, and Percy Liang.
  2018.
\newblock Reinforcement learning on web interfaces using workflow-guided
  exploration.
\newblock \emph{arXiv preprint arXiv:1802.08802}.

\bibitem[{Liu et~al.(2023)Liu, Yu, Zhang, Xu, Lei, Lai, Gu, Ding, Men, Yang,
  Zhang, Deng, Zeng, Du, Zhang, Shen, Zhang, Su, Sun, Huang, Dong, and
  Tang}]{liu2023agentbench}
Xiao Liu, Hao Yu, Hanchen Zhang, Yifan Xu, Xuanyu Lei, Hanyu Lai, Yu~Gu,
  Hangliang Ding, Kaiwen Men, Kejuan Yang, Shudan Zhang, Xiang Deng, Aohan
  Zeng, Zhengxiao Du, Chenhui Zhang, Sheng Shen, Tianjun Zhang, Yu~Su, Huan
  Sun, Minlie Huang, Yuxiao Dong, and Jie Tang. 2023.
\newblock AgentBench: Evaluating LLMs as Agents.
\newblock \emph{arXiv preprint arXiv: 2308.03688}.

\bibitem[{Madaan et~al.(2023)Madaan, Tandon, Gupta, Hallinan, Gao, Wiegreffe,
  Alon, Dziri, Prabhumoye, Yang, Welleck, Majumder, Gupta, Yazdanbakhsh, and
  Clark}]{madaan2023selfrefine}
Aman Madaan, Niket Tandon, Prakhar Gupta, Skyler Hallinan, Luyu Gao, Sarah
  Wiegreffe, Uri Alon, Nouha Dziri, Shrimai Prabhumoye, Yiming Yang, Sean
  Welleck, Bodhisattwa~Prasad Majumder, Shashank Gupta, Amir Yazdanbakhsh, and
  Peter Clark. 2023.
\newblock \href {http://arxiv.org/abs/2303.17651} {Self-Refine: Iterative
  Refinement with Self-Feedback}.

\bibitem[{Pan et~al.(2024)Pan, Zhang, Tomlin, Zhou, Levine, and
  Suhr}]{pan2024autonomous}
Jiayi Pan, Yichi Zhang, Nicholas Tomlin, Yifei Zhou, Sergey Levine, and Alane
  Suhr. 2024.
\newblock \href {http://arxiv.org/abs/2404.06474} {Autonomous Evaluation and
  Refinement of Digital Agents}.

\bibitem[{Prasad et~al.(2023)Prasad, Koller, Hartmann, Clark, Sabharwal,
  Bansal, and Khot}]{prasad2023adapt}
Archiki Prasad, Alexander Koller, Mareike Hartmann, Peter Clark, Ashish
  Sabharwal, Mohit Bansal, and Tushar Khot. 2023.
\newblock ADaPT: As-Needed Decomposition and Planning with Language Models.
\newblock \emph{arXiv}.

\bibitem[{Shinn et~al.(2023)Shinn, Cassano, Berman, Gopinath, Narasimhan, and
  Yao}]{shinn2023reflexion}
Noah Shinn, Federico Cassano, Edward Berman, Ashwin Gopinath, Karthik
  Narasimhan, and Shunyu Yao. 2023.
\newblock \href {http://arxiv.org/abs/2303.11366} {Reflexion: Language Agents
  with Verbal Reinforcement Learning}.

\bibitem[{Shridhar et~al.(2021)Shridhar, Yuan, C\^ot\'e, Bisk, Trischler, and
  Hausknecht}]{ALFWorld20}
Mohit Shridhar, Xingdi Yuan, Marc-Alexandre C\^ot\'e, Yonatan Bisk, Adam
  Trischler, and Matthew Hausknecht. 2021.
\newblock \href {https://arxiv.org/abs/2010.03768} {{ALFWorld: Aligning Text
  and Embodied Environments for Interactive Learning}}.
\newblock In \emph{Proceedings of the International Conference on Learning
  Representations (ICLR)}.

\bibitem[{Song et~al.(2023)Song, Wu, Washington, Sadler, Chao, and
  Su}]{song2023llmplanner}
Chan~Hee Song, Jiaman Wu, Clayton Washington, Brian~M. Sadler, Wei-Lun Chao,
  and Yu~Su. 2023.
\newblock LLM-Planner: Few-Shot Grounded Planning for Embodied Agents with
  Large Language Models.
\newblock In \emph{Proceedings of the IEEE/CVF International Conference on
  Computer Vision (ICCV)}.

\bibitem[{Song et~al.(2024)Song, Yin, Yue, Huang, Li, and Lin}]{song2024trial}
Yifan Song, Da~Yin, Xiang Yue, Jie Huang, Sujian Li, and Bill~Yuchen Lin. 2024.
\newblock Trial and Error: Exploration-Based Trajectory Optimization for LLM
  Agents.
\newblock \emph{arXiv preprint arXiv:2403.02502}.

\bibitem[{Sun et~al.(2023)Sun, Zhuang, Kong, Dai, and
  Zhang}]{sun2023adaplanner}
Haotian Sun, Yuchen Zhuang, Lingkai Kong, Bo~Dai, and Chao Zhang. 2023.
\newblock \href {http://arxiv.org/abs/2305.16653} {AdaPlanner: Adaptive
  Planning from Feedback with Language Models}.

\bibitem[{Wang et~al.(2023{\natexlab{a}})Wang, Xie, Jiang, Mandlekar, Xiao,
  Zhu, Fan, and Anandkumar}]{wang2023voyager}
Guanzhi Wang, Yuqi Xie, Yunfan Jiang, Ajay Mandlekar, Chaowei Xiao, Yuke Zhu,
  Linxi Fan, and Anima Anandkumar. 2023{\natexlab{a}}.
\newblock Voyager: An Open-Ended Embodied Agent with Large Language Models.
\newblock \emph{arXiv preprint arXiv: Arxiv-2305.16291}.

\bibitem[{Wang et~al.(2023{\natexlab{b}})Wang, Li, Chen, Cai, Zhu, Lin, Cao,
  Liu, Liu, and Sui}]{wang2023large}
Peiyi Wang, Lei Li, Liang Chen, Zefan Cai, Dawei Zhu, Binghuai Lin, Yunbo Cao,
  Qi~Liu, Tianyu Liu, and Zhifang Sui. 2023{\natexlab{b}}.
\newblock \href {http://arxiv.org/abs/2305.17926} {Large Language Models are
  not Fair Evaluators}.

\bibitem[{Wang et~al.(2022)Wang, Jansen, C{\^o}t{\'e}, and
  Ammanabrolu}]{wang-etal-2022-scienceworld}
Ruoyao Wang, Peter Jansen, Marc-Alexandre C{\^o}t{\'e}, and Prithviraj
  Ammanabrolu. 2022.
\newblock \href {https://doi.org/10.18653/v1/2022.emnlp-main.775}
  {{S}cience{W}orld: Is your Agent Smarter than a 5th Grader?}
\newblock In \emph{Proceedings of the 2022 Conference on Empirical Methods in
  Natural Language Processing}, pages 11279--11298, Abu Dhabi, United Arab
  Emirates. Association for Computational Linguistics.

\bibitem[{Wang et~al.(2023{\natexlab{c}})Wang, Cai, Chen, Liu, Ma, and
  Liang}]{wang2023describe}
Zihao Wang, Shaofei Cai, Guanzhou Chen, Anji Liu, Xiaojian Ma, and Yitao Liang.
  2023{\natexlab{c}}.
\newblock \href {https://openreview.net/forum?id=KtvPdGb31Z} {Describe,
  Explain, Plan and Select: Interactive Planning with {LLM}s Enables Open-World
  Multi-Task Agents}.
\newblock In \emph{Thirty-seventh Conference on Neural Information Processing
  Systems}.

\bibitem[{Wu et~al.(2023)Wu, Wang, Xu, Lu, and Yan}]{TaPA}
Zhenyu Wu, Ziwei Wang, Xiuwei Xu, Jiwen Lu, and Haibin Yan. 2023.
\newblock Embodied Task Planning with Large Language Models.
\newblock \emph{arXiv preprint arXiv:2305.03716}.

\bibitem[{Xu et~al.(2023)Xu, Peng, Lei, Mukherjee, Liu, and Xu}]{xu2023rewoo}
Binfeng Xu, Zhiyuan Peng, Bowen Lei, Subhabrata Mukherjee, Yuchen Liu, and
  Dongkuan Xu. 2023.
\newblock \href {http://arxiv.org/abs/2305.18323} {ReWOO: Decoupling Reasoning
  from Observations for Efficient Augmented Language Models}.

\bibitem[{Yao et~al.(preprint)Yao, Chen, Yang, and Narasimhan}]{yao2022webshop}
Shunyu Yao, Howard Chen, John Yang, and Karthik Narasimhan. preprint.
\newblock WebShop: Towards Scalable Real-World Web Interaction with Grounded
  Language Agents.
\newblock In \emph{ArXiv}.

\bibitem[{Yao et~al.(2023{\natexlab{a}})Yao, Yu, Zhao, Shafran, Griffiths, Cao,
  and Narasimhan}]{yao2023tree}
Shunyu Yao, Dian Yu, Jeffrey Zhao, Izhak Shafran, Thomas~L. Griffiths, Yuan
  Cao, and Karthik~R Narasimhan. 2023{\natexlab{a}}.
\newblock \href {https://openreview.net/forum?id=5Xc1ecxO1h} {Tree of Thoughts:
  Deliberate Problem Solving with Large Language Models}.
\newblock In \emph{Thirty-seventh Conference on Neural Information Processing
  Systems}.

\bibitem[{Yao et~al.(2023{\natexlab{b}})Yao, Zhao, Yu, Du, Shafran, Narasimhan,
  and Cao}]{yao2023react}
Shunyu Yao, Jeffrey Zhao, Dian Yu, Nan Du, Izhak Shafran, Karthik Narasimhan,
  and Yuan Cao. 2023{\natexlab{b}}.
\newblock {ReAct}: Synergizing Reasoning and Acting in Language Models.
\newblock In \emph{International Conference on Learning Representations
  (ICLR)}.

\bibitem[{Zheng et~al.(2023)Zheng, Chiang, Sheng, Zhuang, Wu, Zhuang, Lin, Li,
  Li, Xing, Zhang, Gonzalez, and Stoica}]{zheng2023judging}
Lianmin Zheng, Wei-Lin Chiang, Ying Sheng, Siyuan Zhuang, Zhanghao Wu, Yonghao
  Zhuang, Zi~Lin, Zhuohan Li, Dacheng Li, Eric Xing, Hao Zhang, Joseph~E.
  Gonzalez, and Ion Stoica. 2023.
\newblock \href {https://openreview.net/forum?id=uccHPGDlao} {Judging
  {LLM}-as-a-Judge with {MT}-Bench and Chatbot Arena}.
\newblock In \emph{Thirty-seventh Conference on Neural Information Processing
  Systems Datasets and Benchmarks Track}.

\bibitem[{Zhou et~al.(2024{\natexlab{a}})Zhou, Yan, Shlapentokh-Rothman, Wang,
  and Wang}]{zhou2024language}
Andy Zhou, Kai Yan, Michal Shlapentokh-Rothman, Haohan Wang, and Yu-Xiong Wang.
  2024{\natexlab{a}}.
\newblock \href {https://openreview.net/forum?id=6LNTSrJjBe} {Language Agent
  Tree Search Unifies Reasoning Acting and Planning in Language Models}.

\bibitem[{Zhou et~al.(2024{\natexlab{b}})Zhou, Xu, Zhu, Zhou, Lo, Sridhar,
  Cheng, Ou, Bisk, Fried, Alon, and Neubig}]{zhou2024webarena}
Shuyan Zhou, Frank~F. Xu, Hao Zhu, Xuhui Zhou, Robert Lo, Abishek Sridhar,
  Xianyi Cheng, Tianyue Ou, Yonatan Bisk, Daniel Fried, Uri Alon, and Graham
  Neubig. 2024{\natexlab{b}}.
\newblock \href {https://openreview.net/forum?id=oKn9c6ytLx} {WebArena: A
  Realistic Web Environment for Building Autonomous Agents}.
\newblock In \emph{The Twelfth International Conference on Learning
  Representations}.

\bibitem[{Zhu et~al.(2023)Zhu, Chen, Tian, Tao, Su, Yang, Huang, Li, Lu, Wang,
  Qiao, Zhang, and Dai}]{zhu2023ghost}
Xizhou Zhu, Yuntao Chen, Hao Tian, Chenxin Tao, Weijie Su, Chenyu Yang, Gao
  Huang, Bin Li, Lewei Lu, Xiaogang Wang, Yu~Qiao, Zhaoxiang Zhang, and Jifeng
  Dai. 2023.
\newblock Ghost in the Minecraft: Generally Capable Agents for Open-World
  Environments via Large Language Models with Text-based Knowledge and Memory.
\newblock \emph{arXiv preprint arXiv:2305.17144}.

\bibitem[{Zhuang et~al.(2024)Zhuang, Chen, Yu, Mitra, Bursztyn, Rossi, Sarkhel,
  and Zhang}]{zhuang2024toolchain}
Yuchen Zhuang, Xiang Chen, Tong Yu, Saayan Mitra, Victor Bursztyn, Ryan~A.
  Rossi, Somdeb Sarkhel, and Chao Zhang. 2024.
\newblock \href {https://openreview.net/forum?id=B6pQxqUcT8} {ToolChain*:
  Efficient Action Space Navigation in Large Language Models with A* Search}.
\newblock In \emph{The Twelfth International Conference on Learning
  Representations}.

\end{thebibliography}
\section*{Appendix}
\label{appendix}

\subsection*{Prompt for Plan Generation (\(G_{\text{plan}}\))}
Imagine that you are imitating humans doing a task on a website step by step. You can click an element with the mouse, scroll up or down, go to a certain URL or go back to previous page, or type some text with the keyboard (e.g., click(), scroll(), goto(), go\_back(), and type() functions in playwright). One step means one operation within any of the mentioned actions.

You are within a sandbox and only have access to the following websites to work with:

\begin{itemize}

\item An online shopping website (OneStopShop): \{webarena\_root\}:7770

\item An e-commerce management website (Magento): \{webarena\_root\}:7780/admin

\item A Reddit website (Postmill): \{webarena\_root\}:9999

\item A GitLab website: \{webarena\_root\}:8023

\item A map website (OpenStreetMap): \url{http://ec2-3-131-244-37.us-east-2.compute.amazonaws.com:3000}

\item A Wikipedia website: \url{http://ec2-3-131-244-37.us-east-2.compute.amazonaws.com:8888/wikipedia\_en\_all\_maxi\_2022-05/A/User:The\_other\_Kiwix\_guy/Landing}

\end{itemize}

\textbf{Notes:}

\begin{enumerate}

\item If you want to use the search function, you don't need to click on the search bar. You can directly use ``type [element\_id] [things\_to\_type]'', and generally afterwards, you don't need to click the search button (by default, the command contains an ENTER at the end).

\item You can assume that you have signed in to your account (we have set up the cookies, so login is not needed).

\end{enumerate}

The website that you will be working with is:

\{WEBSITE INTRO\}

Please follow these specific instructions to solve tasks:

\{INSTRUCTION\}

Here is a more detailed description of the starting screen: 

\{STARTING SCREEN DESCRIPTION\}

Now, based on the information above, what should be the steps to achieve the following goal (please give me a list of textual description of playwright actions, starting with `List'):

\{TASK\}

For your reference, here are some experiences from previous failed trials (please consider the following information to generate a better plan):

\{FAILED PLAN\}

Past experience:

\{HISTORY\}

To be successful in generating a new plan, you need to provide a list (1, 2, 3, ...), in which each item is a natural language description of one playwright action that is necessary to complete the task (e.g., click on the `Account' button; scroll down; use the search bar to search for iPhone 13). You should use the information from the past experiences to save unnecessary steps!

\subsection*{Prompt for Action Generation (\(G_{\text{action}}\))}
I am in a sandbox and only have access to the following websites (i.e., no access to external website like www.reddit.com):

\begin{itemize}

\item An online shopping website (OneStopShop): \{webarena\_root\}:7770

\item An e-commerce management website (Magento): \{webarena\_root\}:7780/admin

\item A Reddit website (Postmill): \{webarena\_root\}:9999

\item A GitLab website: \{webarena\_root\}:8023

\item A map website (OpenStreetMap): \url{http://ec2-3-131-244-37.us-east-2.compute.amazonaws.com:3000}

\item A Wikipedia website: \url{http://ec2-3-131-244-37.us-east-2.compute.amazonaws.com:8888/wikipedia\_en\_all\_maxi\_2022-05/A/User:The\_other\_Kiwix\_guy/Landing}

\end{itemize}

Now I'm trying to complete a task on a website.

The task is: 

\{TASK\}

The plan to complete this task is:

\{PLAN\}

I have executed the following actions: 

\{HISTORY\}

And now I'm at this step:
\{STEP\}

Here is the screen I am looking at:

\{OBS\}

I have taken down the following notes:

\{NOTES\}

What should be the next action to complete this step in my plan (only give one action)?

\textbf{Note:}

\begin{itemize}

\item If the next action is to click, please indicate the element id in [] (format: click [element\_id]).

\item If the next action is to scroll, please indicate the direction in [] (format: scroll [up or down]).

\item If you need to navigate to a URL, please indicate the URL in [] (format: goto [url]).

\item If you need to go back to the previous page, please use go\_back.

\item If the next action is to type, please indicate both element id and the things to type in [] (format: type [element\_id] [things to type]).

\item If you want to note down something, use this format: note\_down [things to note down].

\end{itemize}

The next action is:

\subsection*{Prompt for Objective Alignment (\(G_{\text{align}}\))}
Imagine that you are imitating humans doing a task on a website step by step. 

You are currently working on this step:

\{STEP\}.

The step above is one of the steps in the following plan:

\{PLAN\}.

From Screen 1, you executed an action and then arrived at Screen 2. 

The action you executed was: 

\{ACTION\}. 

Screen 1: 

\{OBS1\}. 

Screen 2: 

\{OBS2\}. 

Now describe what this action is about in one sentence, starting with `The action is to'.

Does this action align with the goal of the following step (i.e., are we moving towards the right direction; Answer YES or NO)?  

\{STEP\}

\subsection*{Prompt for Task / Subtask Completion Evaluation (\(G_{\text{completed}}\))}

Imagine that you are imitating humans doing a task on a website step by step.

You are asked to solve the following task:

\{TASK\}

You made the following plan to solve it:

\{PLAN\}

To reach the current screen, you have previously executed the following actions:

\{HISTORY\}

You have taken down a few notes after each action as follows:

\{NOTES\}

And here is the accessibility tree of the current screen you are looking at:

\{OBS\}

Look at the screen, the task, and the actions you executed, and think thoroughly, is the task completed now?

If the task is completed, answer YES. 

If the task is not yet completed (meaning further actions are yet to be executed), answer NO.

\subsection*{Prompt for Answer Delivery (\(G_{\text{answer}}\))}
Imagine that you are imitating humans doing a task on a website step by step.

You are asked to solve the following task:

\{TASK\}

To reach the current screen, you have previously executed the following actions:

\{HISTORY\}

You have taken down the following notes (to help you answer the question eventually) after each action:

\{NOTES\}

And here is the accessibility tree of the current screen you are looking at:

\{OBS\}

Based on the above information, answer the question in the task (starting with \#\#\#Answer).

\subsection*{Prompt for Element Mapping (\(G_{\text{map}}\))}
I want to interact with an element with element id:
\{element\_id\} in the following screen:

\{OBS1\}

Now if I want to click on the same element in the following screen, what should be the element id now?

\{OBS2\}

New element id is:

\end{document}